\documentclass[11pt]{article}
\usepackage{adjustbox}
\usepackage{acl}
\usepackage{times}
\usepackage{graphicx}
\usepackage{subcaption}
\usepackage{latexsym}
\usepackage{booktabs}
\usepackage{multirow}
\usepackage{amssymb}
\usepackage{algorithm}

\usepackage{algpseudocode}
\usepackage{tabularx}
\usepackage[T1]{fontenc}
\usepackage{amsmath}
\usepackage[utf8]{inputenc}

\usepackage{microtype}

\usepackage{inconsolata}

\usepackage{booktabs}
\usepackage[table]{xcolor} 
\usepackage{geometry}

\usepackage{enumitem}

\usepackage{graphicx}

\newcommand{\ours}{\textbf{\texttt{ESTP}}}

\title{When Entropy Is Not Enough:  Reclaiming Lost Semantics \\ in  LLM Output Length Prediction}

\author{
\textbf{Feiyang Ren}\textsuperscript{\rm 1}\footnotemark[1]~,  
\textbf{Shengtao Wen}\textsuperscript{\rm 1}\thanks{\;Equal Contribution.}~,
\textbf{Yu Tian}\textsuperscript{\rm 2},
\textbf{Yuanning Cui}\textsuperscript{\rm 3} \\
\textbf{Xiaofeng Cao}\textsuperscript{\rm 4}\textbf{,}
\textbf{Jiangchao Yao}\textsuperscript{\rm 5}\textbf{,} 
\textbf{Xu Chen}\textsuperscript{\rm 6}\textbf{,}
\textbf{Xiang Chen}\textsuperscript{\rm 1}\thanks{\;Corresponding author.}
\\
\textsuperscript{\rm 1}
Nanjing University of Aeronautics and Astronautics \\
  \textsuperscript{\rm 2}360 Security Capability Center \\
  \textsuperscript{\rm 3}Nanjing University of Information Science and Technology \\
  \textsuperscript{\rm 4}Tongji University
  \textsuperscript{\rm 5}Shanghai Jiao Tong University
  \textsuperscript{\rm 6}Alibaba Group \\
    { \texttt{\{feiyang\_ren,xiang\_chen\}@nuaa.edu.cn}}\\
}

\begin{document}

\maketitle

\begin{abstract}
Efficient LLM serving is often bottlenecked by the need to pad sequences to a fixed maximum length, and this wastes compute and degrades throughput. Output length prediction enables length-aware scheduling and reduces overhead. Existing methods, including entropy-guided token pooling, rely on token-level entropy but overlook semantic differences between tokens, causing them to underweight important tokens and overemphasize less informative tokens. We introduce \textbf{ESTP}(\textbf{E}ntropy-and-\textbf{S}emantic \textbf{T}oken \textbf{P}ooling), a lightweight framework that combines entropy representations with attention-based semantic representations. By deriving both representations directly during prefill, ESTP captures uncertainty and semantic importance while incurring negligible additional inference latency and memory overhead compared with methods using external auxiliary models. 
On the ForeLen benchmark, ESTP outperforms baseline methods, achieves better prediction accuracy and lower prediction errors across most scenarios. 
When integrated with a length-aware scheduler in end-to-end system tests, it further improves overall throughput and reduces the padding ratio.
\end{abstract}

\section{Introduction}

Efficient serving of Large Language Models (LLMs) remains a critical system-level challenge. Although continuous batching, PagedAttention~\cite{Kwon}, PowerAttention~\cite{powerattention}, and SARATHI~\cite{SARATHI} have improved throughput, the barrel effect, which wastes computation by padding shorter sequences to the longest sequence in a batch~\cite{barrel-effect}, remains a fundamental bottleneck. 
\begin{figure}[t!]
    \centering
    \begin{subfigure}[b]{1.0\linewidth}
        \centering
        \includegraphics[width=1.0\linewidth]{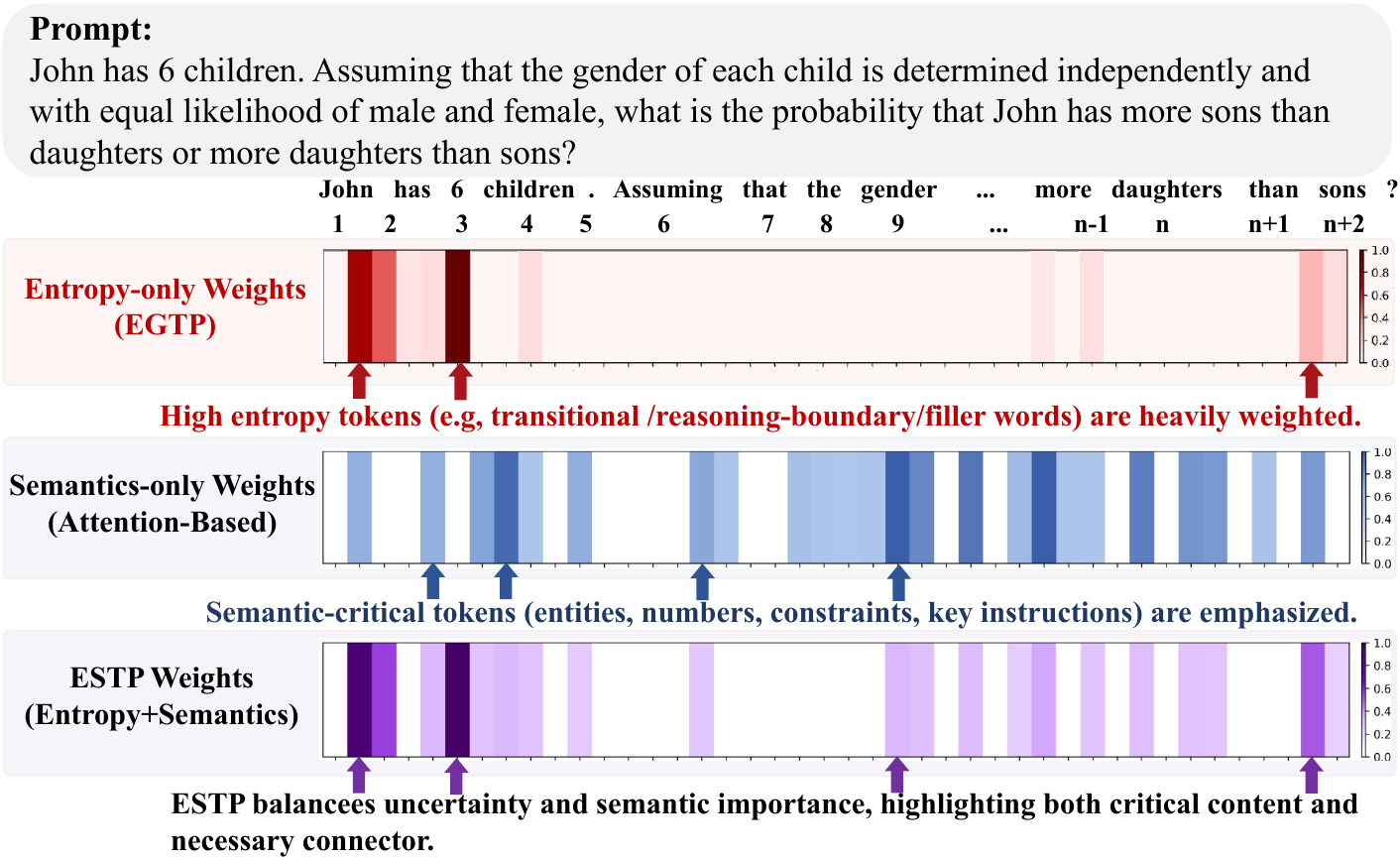}
        \caption{Balanced Token Weighting with ESTP}
        \label{fig:pic-a}
    \end{subfigure}
    \begin{subfigure}[b]{1.0\linewidth}
        \centering
        \includegraphics[width=1.0\linewidth]{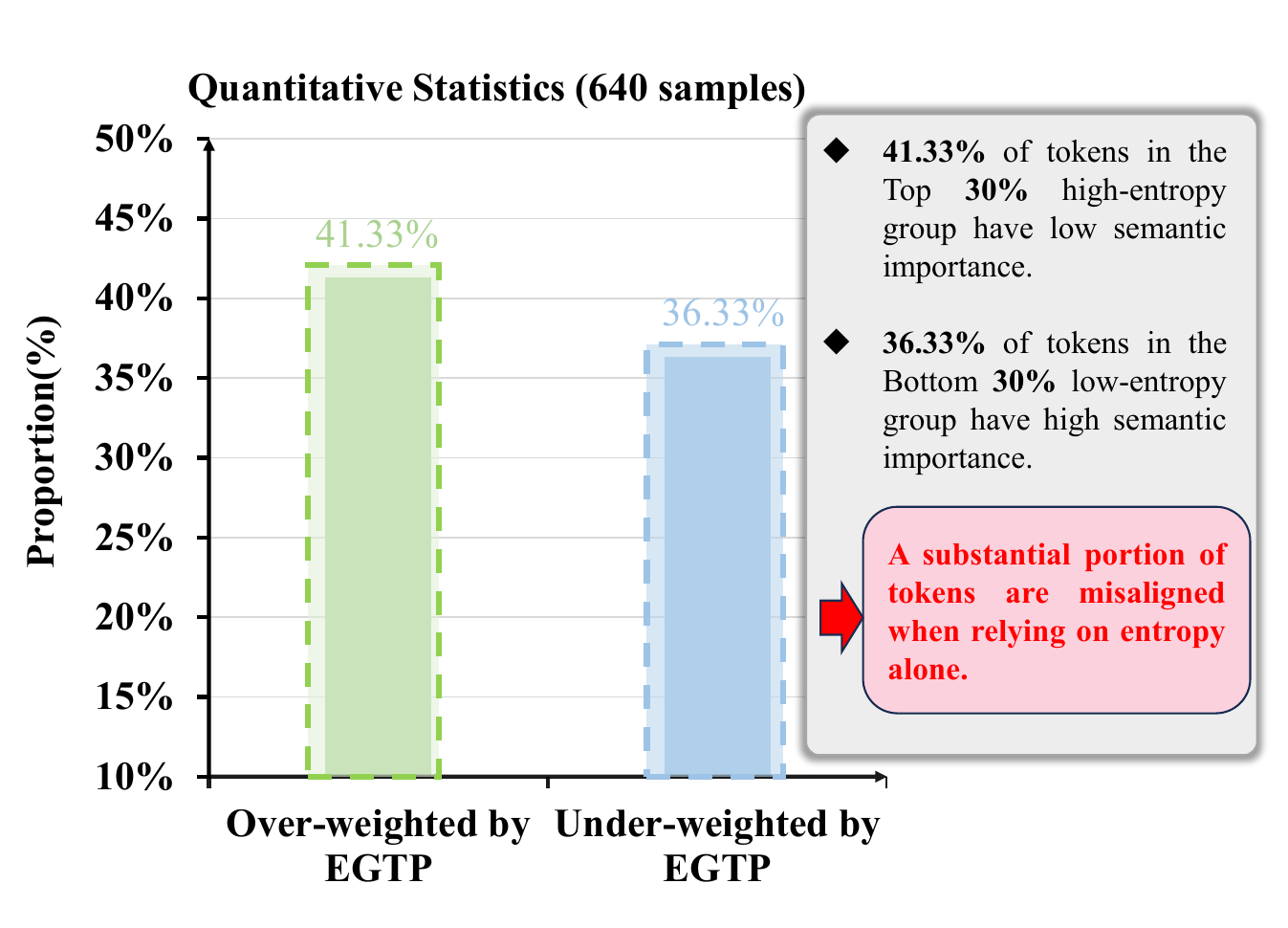}
        \caption{Quantitative Results of Entropy-Semantic Misalignment}
        \label{fig:pic-b}
    \end{subfigure}
    \caption{\textbf{Limitations of Entropy-Only Token Importance Assessment.} (a) Comparison of Token-Level Weighting Strategies: Entropy-Only, Semantics-Only, and ESTP. (b) Quantitative Analysis of Token Misalignment Between Entropy and Semantic Importance.}
    \label{fig:pic}
\end{figure}
This overhead is especially severe in workloads with high length variance, including long-context reasoning~\cite{longtext,longtext1} and dynamic Reinforcement Learning (RL) sampling~\cite{RL1,RL2}.

Accurate pre-generation output length prediction enables length-aware scheduling, reducing padding waste and improving hardware utilization. Existing prediction methods range from costly external auxiliary models, such as DistilBERT-based classifiers~\cite{Jin, baseline1-2}, to LLM-intrinsic approaches such as TRAIL~\cite{TRAIL}. Recently, EGTP~\cite{entropy} achieved state-of-the-art performance by reusing LLM on-the-fly activations and token-level entropy for weighted pooling with negligible overhead. However, as illustrated in Figure~\ref{fig:pic-a}, relying solely on entropy introduces a systematic blind spot: entropy captures generation uncertainty rather than semantic importance. Core content tokens in a prompt, including key entities, constraints, and instructions, are often processed with high confidence (low entropy), causing them to be under-weighted. In contrast, high-entropy transitional or filler tokens may be over-amplified, leading to substantial semantic signal loss during prediction.

We quantitatively verify this semantic misalignment through a controlled diagnostic on 640 RL-sampled instances. For each token, we compute both its entropy and its semantic importance, and we use the debiased attention weights from the LLM prefill phase as a useful proxy for the latter. The results reveal a substantial mismatch between the two measures in the analyzed samples. As illustrated in Figure~\ref{fig:pic-b}, when tokens are split at the 70th percentile, a substantial fraction of high-entropy tokens carry little semantic importance, and many low-entropy tokens are semantically critical. This suggests that entropy alone may not fully capture token relevance, leaving the predictor vulnerable in complex reasoning where prompt semantics strongly influence output length.

\begin{figure*}[ht]
    \centering
    \includegraphics[width=1\textwidth]{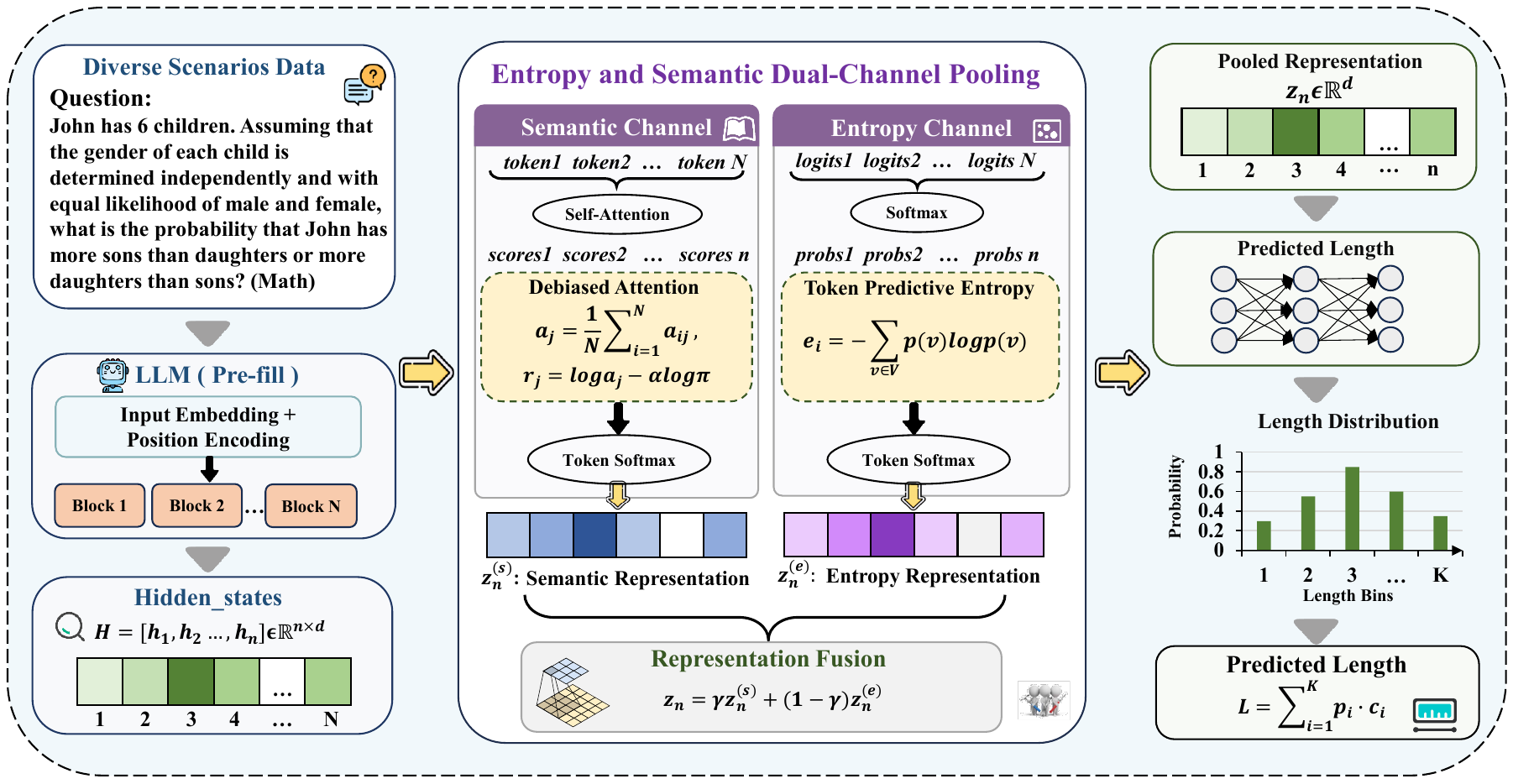} 
    \caption{Overview of the \ours~ framework. The framework consists of three main components: (1) \textbf{Construction of Hidden States}, which extracts hidden state representations from diverse scenarios including reasoning, long-sequence, and RL tasks; (2) \textbf{Semantic and Entropy Dual-Channel Pooling}, which computes attention-based semantic weights and aggregates entropy values to identify high-entropy frequent tokens (e.g., logical connectives, suppositional words, inference steps), yielding combined representations $z_n$; and (3) \textbf{Lightweight Prediction}, which applies softmax-normalized weights to hidden states and feeds the weighted representation into an MLP predictor to estimate the output length with a distributional CRPS loss and a continuous SmoothL1 regression loss.}
    \label{fig:method}
\end{figure*}

To recover these lost semantics, we propose {\ours} (\textbf{E}ntropy-and-\textbf{S}emantic \textbf{T}oken \textbf{P}ooling), a lightweight prediction framework shown in Figure~\ref{fig:method}. {\ours} separately derives a semantic-weighted feature using debiased attention scores and an entropy-weighted feature using token predictive entropy, and then linearly fuses the two features. The resulting aggregated representation captures both uncertainty signals and semantic information. During training, we use an ordinal length-distribution prediction head to model the ordered structure of length bins. We replace EGTP’s original combined loss with discrete CRPS and normalized Smooth L1 to reduce sensitivity to heavy-tailed targets and extreme errors. The resulting model estimates the expected length more accurately and produces probability distributions that are better suited to tail-risk estimation and length-aware scheduling. Because {\ours} reuses prefill features, it adds little inference latency or memory overhead compared with methods that use external auxiliary models.

We evaluate {\ours} on ForeLen~\cite{entropy}, a custom dataset spanning three challenging scenarios (long-sequence generation, complex reasoning, and dynamic RL sampling), using four LLM backbones (Qwen2.5-3B/7B and Llama3.2-1B/3B). We also compare the results of {\ours} with the strongest baseline under the extreme long‑CoT regime. Our main contributions are summarized as follows.
\begin{itemize}
    \item We empirically verify a systematic mismatch between entropy and semantic importance. Tokens that carry critical semantic content often exhibit high prediction confidence, and this leads to insufficient weight assignment. This finding challenges the common assumption that uncertainty alone can serve as a reliable indicator of token relevance.
    \item We propose {\ours}. It combines entropy-based and attention-based semantic pooling with an ordinal length-distribution prediction head. By balancing uncertainty and semantic importance, it recovers the semantic signal that entropy underweights, and it does this at negligible additional inference cost compared with methods using external auxiliary models.
    \item Extensive evaluations show that {\ours} offers a practical improvement for length prediction. Across a range of models and most scenarios on ForeLen benchmark, our method generally reduces prediction error (MAE), improves binning accuracy, and further enhances throughput while reducing the padding ratio on end-to-end tasks.
\end{itemize}

\section{Related Work}
\label{sec:related_work}

\subsection{LLM Serving Optimization}
Existing LLM serving optimizations, including continuous batching~\cite{Kwon}, batch prompting~\cite{batch1}, and PagedAttention~\cite{Kwon}, improve throughput by dynamically managing requests. Recent studies have further advanced the field through improved scheduling strategies, memory management, and parallel computing techniques. For example, SARATHI~\cite{SARATHI} improves decoding throughput with chunked prefill and decode piggybacking. However, these methods do not address the ``barrel effect'', which refers to the computational waste caused by padding shorter sequences to match the longest sequence in a batch~\cite{barrel-effect}. Our work takes a complementary approach by reusing hidden states generated internally by LLMs. This enables accurate output length prediction and length-aware scheduling, reducing padding overhead while complementing existing system-level optimizations.

\subsection{LLM Output Length Prediction}

Conventional approaches to LLM output length prediction often rely on external auxiliary models, such as DistilBERT-based classifiers~\cite{Jin,baseline1-2,Hu}. These models only process input prompts and require substantial computational resources and training time. To overcome these limitations, a series of lighter‑weight methods based on LLM internal activations have emerged. TRAIL~\cite{TRAIL} introduced a prediction method based on the target LLM itself, achieving low overhead and accurate predictions. ProD~\cite{robust} methods construct robust supervision signals via multiple generations for the same prompt, and leverage LLM hidden states to achieve more robust output‑length prediction. Inspired by studies connecting model internals, such as token entropy, with generation characteristics~\cite{Li}, EGTP (Entropy-Guided Token Pooling)~\cite{entropy} directly reused LLM on-the-fly activations and token entropy for accurate static prediction with negligible overhead. This approach significantly reduces training costs. However, while this lightweight paradigm enables efficient inference and accurate prediction, relying solely on entropy weighting can lead to semantic information loss. Our framework addresses this limitation by combining entropy with semantic information to construct an improved weighted pooling mechanism.

\section{Preliminaries}
\label{sec:preliminaries}

\begin{figure}[t]
    \centering
    \begin{subfigure}[b]{0.95\linewidth}
        \centering
        \includegraphics[width=0.95\linewidth]{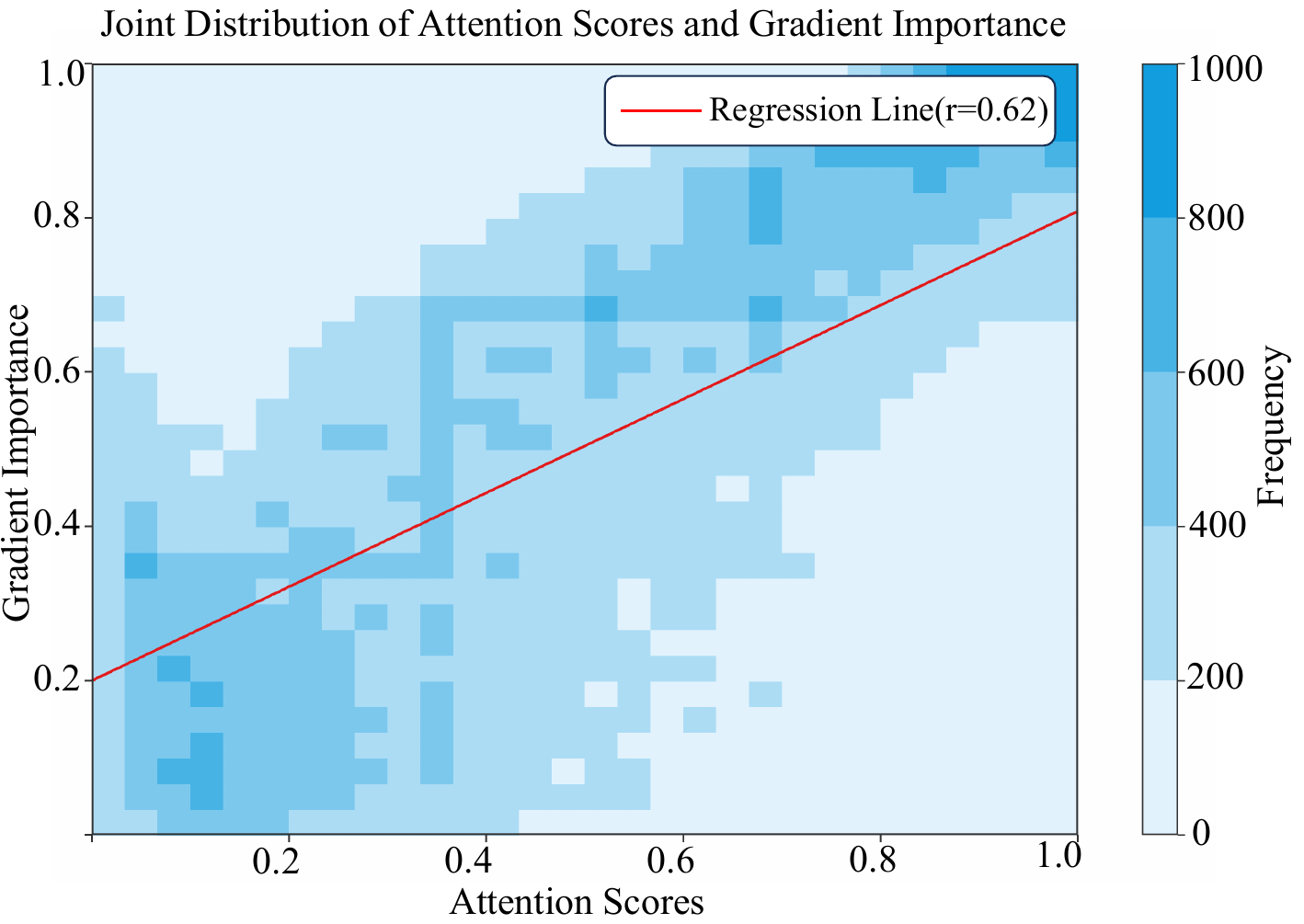}
        \label{fig:relitu}
    \end{subfigure}
    \caption{The joint distribution of Attentions Scores and Gradient Importance, with the regression line confirming a positive correlation (r=0.62).}
    \label{fig:relitu}
\end{figure}

\paragraph{Inherent Limitations of EGTP Mechanisms.}
During LLM inference, token entropy reflects generation uncertainty~\cite{TECP,Beyond-Semantic-Entropy,entropy1,entropy2} and has been shown to correlate positively with output length prediction, which motivates entropy-weighted pooling for lightweight length prediction~\cite{entropy}. But relying only on entropy has a key limitation. High-entropy tokens reflect model uncertainty, and this mechanism dilutes the contribution of semantically core tokens. As a result, critical information gets discarded, irrelevant tokens receive excessive weight, and prediction performance ultimately degrades.

\begin{table}[t]
\centering
\resizebox{0.45\textwidth}{!}{
\begin{tabular}{lcccc}
\toprule
\textbf{Scenario} & \textbf{P@30} &
\textbf{R@30} & \textbf{P@50} & \textbf{R@50}\\
\midrule
RL & 69\% & 43\% & 65\% &73\% \\
 Reasoning  & 66\% & 49\% & 63\% &74\% \\
 Average  & 68\% & 46\% & 64\% &74\% \\
\bottomrule
\end{tabular}
}
\caption{Attention-Semantic Alignment.}
\label{tab:yuyi}
\end{table}

\paragraph{Motivation.}
Length prediction needs an understanding of semantic structure. A length predictor generally needs to identify key entities and core instructions in the query. These signals often show up in tokens that have low entropy but high attention weights. In Transformers, attention weights offer a lightweight signal indicating how strongly tokens contribute to contextual representation construction~\cite{attn13}, such as the use of attention value vectors to capture sentence-level semantics~\cite{attn-semantic2} and the temperature-controlled Softmax sharpening approach that directs model focus toward the most relevant tokens~\cite{attn-semantic1}. And prior work has shown them to be a useful proxy for semantic importance in various settings~\cite{attn1,attn2,attn3,attn4}. To assess whether debiased attention scores reflect human judgments of semantic importance, we randomly sample 120 prompts from each of the RL and Reasoning settings. Three trained annotators independently rate each token on a four-point scale from 0 to 3 using a shared set of annotation guidelines, and tokens scored 2 or 3 are considered important. We evaluate the top 30\% and 50\% of tokens ranked by debiased attention using Precision and Recall. As shown in Table~\ref{tab:yuyi}, expanding the selection from the top 30\% to the top 50\% substantially increases recall, with only a modest decrease in precision, in both the RL and Reasoning settings. 
Based on these results, we use debiased attention as a lightweight proxy for token-level semantic importance rather than treating it as a direct measure.

To better understand the role of token-level attention for length prediction, we measure each token’s task-specific contribution with a gradient-based attribution method~\cite{guiyin,guiyin1,guiyin2}. As shown in Figure~\ref{fig:relitu}, debiased attention scores show a strong and highly consistent association with gradient importance across 5$k$ samples randomly selected from the ForeLen dataset. This association suggests that tokens with higher debiased attention scores tend to carry greater importance for length prediction, supporting the design of ESTP.
We use 640 RL samples to verify the limitations of entropy-only weighting. From LLM activations, we compute token entropy and attention-based semantic scores. We classify tokens into high- and low-entropy and high- and low-semantic-importance groups, then calculate the misalignment ratio (Figure~\ref{fig:pic-b}). Over 40\% of the top 30\% high-entropy tokens have low semantic importance, showing that entropy-only pooling may overweight tokens with relatively low attention-based semantic scores while underweighting core prompt information. Ablation studies confirm semantic features contribute more than entropy alone. Over 35\% of the bottom 30\% low-entropy tokens also carry high semantic value, showing that entropy-only weighting downplays semantically important tokens. To address this issue, we add semantic features as a complementary signal. {\ours} fuses entropy and semantic features, balancing uncertainty and importance to achieve more accurate LLM output length prediction.

\section{Method}
\label{sec:method}

\subsection{Prefill Feature Extraction and Empirical Position-Prior Debiasing}
\label{subsec:Prefill Feature}

We use an attention-based semantic weighting strategy and mask padding positions so that only valid tokens contribute to the token weights.
For attention head $d$ in the final transformer layer, the incoming attention score of token $j$ in sample n is defined as the average attention it receives from all valid query positions:
\begin{equation}
\label{eq:2-0}
a_{j,d,n} =\frac{1}{N_{n} }\sum_{i=1}^{N_{n} } A_{i,j,d,n}  ,
\end{equation}
where $N_{n}$ is the input prompt length, $A_{i,j,d}$ denotes the attention weight from token $i$ to token $j$.

However, raw incoming attention in Causal LMs is systematically influenced by token position, because earlier tokens can be attended to by more subsequent queries. To isolate token-specific evidence, we estimate the empirical prior $\pi$ using the training split. Specifically, for token $j$ in sample n, we select the corresponding position prior according to its input-length bucket $\ell (n)$ and relative-position bin $b(j,N_{n})$. We then subtract the scaled log prior from the log incoming-attention score to obtain the debiased attention score and introduce a parameter $\alpha$ to control the strength of prior correction. This correction substantially reduces the correlation between token scores and the causal position baseline. We provide a detailed analysis in Appendix~\ref{app:prior}. Finally, we standardize the debiased scores within each sample and attention head.
\begin{equation}
\label{eq:2-0-1}
r_{j,d,n}=loga_{j,d,n}-\alpha log(\pi _{d,\ell(n) ,b(j,N_{n}) }  )  ,
\end{equation}
For token entropy computation, we follow the calculation scheme introduced in EGTP. We first derive the entropy value $e_i$ of each individual token. Concretely, given the hidden state $h_i$ mapped from input token $x_i$, entropy is quantified based on the next-token probability distribution $P(v|x_{<i})$ across the entire vocabulary $V$ :
\begin{equation}
\label{eq:2-0-2}
e_{i} = -\sum_{v\in V }P(v|x_{<i})logP(v|x_{<i}),
\end{equation}

\subsection{Head-Gated Dual-Channel Pooling and Joint Weight Fusion}
\label{subsec:joint}

First, we combine the debiased per-head attention scores using learnable global head weights: a temperature-controlled Softmax produces the normalized head weights, and their weighted sum represents a semantic score for each token in sample n. The semantic scores are standardized within each sample and normalized over valid tokens to obtain the semantic pooling distribution:
\begin{equation}
\label{eq:2-1-1}
z^{(s)}_{n} =\sum_{j=1}^{N_{n} }  w^{(s)}_{n,j} h_{n,j} ,
\end{equation}
where $w^{(s)}_{n,j}$ denotes the normalized semantic weight assigned to the $j$-th token in sample n, $h_{n,j}$ denotes the last-layer hidden state of the LLM for token $j$ in sample n.

Similarly, we standardize and pool the entropy signal through a separate channel to obtain the entropy pooling distribution $z^{(e)}_{n}$. The two channels are fused after pooling so that their token rankings remain separately normalized. Then we introduce a hyperparameter $\gamma$ to combine the semantic and entropy pooled representations, the fused representation $z_{n}$ is passed to the length distribution head.
\begin{equation}
\label{eq:2-1-2}
z_{n}=\gamma z^{(s)}_{n} + (1-\gamma )z^{(e)} _{n} 
\end{equation}

\begin{table*}[t]
\centering
\small
\setlength{\tabcolsep}{7pt}
\renewcommand{\arraystretch}{1.15}

\resizebox{0.92\textwidth}{!}{
\begin{tabular}{llcccccc}
\toprule

\multirow{2}{*}{\textbf{Model}} 
& \multirow{2}{*}{\textbf{Scenario}} 
& \multicolumn{6}{c}{\textbf{Prediction Method}} \\

\cmidrule(lr){3-8}

& 
& SSJF-Reg 
& SSJF-MC 
& LTR-C 
& TRAIL 
& EGTP 
&  \cellcolor{blue!8}\textbf{ESTP} \\

\midrule

\multirow{4}{*}{\textbf{Qwen2.5-3B}}
& LongSeq   
& 246.76 & 318.71 & 164.64 & 140.31 
& \underline{97.50} 
&  \cellcolor{blue!8}\textbf{89.47} \\

& Reasoning 
& 322.77 & 242.26 & 287.77 & \underline{138.01} 
& 143.11
& 
 \cellcolor{blue!8}\textbf{124.9} \\

& RL        
& 120.37 & 206.42 & \underline{96.65} & 153.72 
& 99.78 
&  \cellcolor{blue!8}\textbf{73.38} \\
\rowcolor{gray!8}
& \textbf{Avg} 
& 229.97 & 255.80 & 183.02 & 144.01 
& \underline{113.46} 
&  \cellcolor{blue!8}\textbf{95.92} \\

\midrule

\multirow{4}{*}{\textbf{Qwen2.5-7B}}

& LongSeq   
& 206.11 & 346.96 & 169.28 & 124.05 
& \underline{88.29} 
&  \cellcolor{blue!8}\textbf{85.13} \\

& Reasoning 
& 299.45 & 284.53 & 256.72 & \underline{133.42} 
& 135.32 
&  \cellcolor{blue!8}\textbf{126.81} \\

& RL        
& 123.56 & 222.65 & 99.70 & 175.55 
& \underline{95.24} 
&  \cellcolor{blue!8}\textbf{74.00} \\
\rowcolor{gray!8}
& \textbf{Avg} 
& 209.71 & 284.71 & 175.23 & 144.34
& \underline{106.28} 
&  \cellcolor{blue!8}\textbf{95.31} \\

\midrule

\multirow{4}{*}{\textbf{Llama3.2-1B}}

& LongSeq   
& 442.27 & 199.40 & 402.98 & 156.30 
& \underline{82.64} 
&  \cellcolor{blue!8}\textbf{72.69} \\

& Reasoning 
& 320.31 & 203.96 & 299.10 & 144.96 
& \underline{138.04} 
&  \cellcolor{blue!8}\textbf{133.71} \\

& RL        
& \underline{88.43} & 192.75 & 122.37 & 172.75 
& 108.53 
&  \cellcolor{blue!8}\textbf{71.89} \\
\rowcolor{gray!8}
& \textbf{Avg} 
& 283.67 & 198.70 & 274.82 & 158.00
& \underline{109.74} 
&  \cellcolor{blue!8}\textbf{92.76} \\

\midrule

\multirow{4}{*}{\textbf{Llama3.2-3B}}

& LongSeq   
& 431.21 & 210.73 & 395.66 & 144.36 
& \underline{122.88} 
&  \cellcolor{blue!8}\textbf{108.82} \\

& Reasoning 
& 359.37 & 200.60 & 334.44 & 176.26 
& \underline{110.41} 
&  \cellcolor{blue!8}\textbf{100.56} \\

& RL        
& \underline{108.65} & 207.20 & 123.48 & 160.63 
& 114.31 
&  \cellcolor{blue!8}\textbf{91.57} \\
\rowcolor{gray!8}
& \textbf{Avg} 
& 299.74 & 206.18 & 284.53 & 160.42 
& \underline{115.87} 
&  \cellcolor{blue!8}\textbf{100.32} \\

\bottomrule
\end{tabular}
}
\caption{
Comparison of MAE across different length prediction methods 
(\textbf{lower is better}). 
Best results are highlighted in \textbf{bold}, and second-best results are \underline{underlined}. 
ESTP consistently achieves lower average prediction error across all backbone LLMs.
}

\label{tab:main_exp}
\end{table*}

\subsection{Ordinal Length Distribution Modeling and Robust Optimization}
\label{subsec:predict}

To address the limitation that the standard MSE loss is highly sensitive to outliers and heavy-tailed distributions, EGTP~\cite{entropy} employs a joint loss function that combines cross-entropy loss and mean squared error loss. However, the soft label relies on a manually specified decay function and only indirectly captures the order and distance between length bins. Meanwhile, MSE applies a quadratic penalty to large errors, making it sensitive to extreme samples in long-tailed length distributions. We therefore replace the original loss with CRPS~\cite{crps} and normalized SmoothL1~\cite{smoothl1}.

We partition the continuous length space into $K$ predefined bins. Given the input representation $z_{n}$, the model predicts a probability distribution $\hat{p}$ over $K$ ordered length bins. Let $c_i$ denote the center of the $i$-th bin, calculation formula defined as:
\begin{equation}
\label{eq:3-2}
\hat{y} =\sum_{i=1}^{K} \hat{p_i}\cdot c_{i}  
\end{equation}
To explicitly model the order and distance between length bins, we use the discrete continuous ranked probability score (discrete CRPS) to supervise the predicted length distribution:
\begin{equation}
\label{eq:3-3}
\mathcal{L}_{\mathrm{CRPS}}=\sum_{k=1}^{K-1}(C_{k}-C^{*}_{k})^{2} \Delta b_{k} 
\end{equation}
Here, $\Delta b_{k}$ is the width of the $k$-th interval, $C_{k}$ is the predicted cumulative distribution, and $C^{*}_{k}$ is the stepwise cumulative distribution corresponding to the ground-truth length bin $i$.

Ultimately, the model is optimized via a combined loss function integrating discrete CRPS and SmoothL1 losses, with hyperparameter $\lambda $ balancing the two components:
\begin{equation}
\label{eq:3-4}
\mathcal{L}=\mathcal{L}_{CRPS}+ \lambda\mathcal{L}_{SmoothL1}
\end{equation}
Here, $\mathcal{L}_{CRPS}$ learns the ordered length distribution, while normalized $\mathcal{L}_{SmoothL1}$ directly constrains the continuous length prediction and reduces the influence of extreme errors during training. Detailed loss function equations in appendix~\ref{app:smooth}.

\section{Experiments}
\label{sec:experiments}

\subsection{Experimental Setup}

\paragraph{Dataset.}
To evaluate the performance of our \ours~in complex and realistic scenarios, we conduct extensive experiments on \textbf{ForeLen} ~\cite{entropy}. (1) Long-Sequence and Complex Reasoning, prompts for this scenario are sourced from three well-established benchmark datasets: LongBench~\cite{LongBench}, ZeroSCROLLS~\cite{Zeroscrolls} and IFEval~\cite{IFEval}. (2) Dynamic RL Sampling, prompts are selected from six widely math and code reasoning datasets: CRUXEval~\cite{CRUXEval}, GSM8K~\cite{GSM8K}, LiveCodeBench~\cite{Livecodebench}, MATH~\cite{Math}, MBPP~\cite{MBPP} and MMLU-STEM~\cite{MMLU-STEM}. Dataset split details in appendix~\ref{app:dataset split}.

\paragraph{Metrics.}
To quantify prediction deviation, we use MAE as the main metric and compare RMSE across methods. We bin output lengths by the RL dataset distribution and compute per-bin accuracy. We measure inference efficiency by average runtime and GPU memory on fixed samples. To evaluate the end-to-end system performance, we use Throughput, JCT(Job Completion Time) and Padding Ratio. Detailed metric calculations are provided in the appendix~\ref{app:metrics}.

\paragraph{Models and Baselines.}
To evaluate the effectiveness of \ours, we conduct extensive experiments on four models: Qwen2.5-3B/7B~\cite{qwen2.5} and Llama3.2-1B/3B~\cite{llama}. For each architecture, we compare six methods: SSJF-Reg~\cite{baseline1-2}, SSJF-MC~\cite{baseline1-2}, TRAIL~\cite{baseline6}, LTR-C~\cite{baseline7}, EGTP~\cite{entropy}, and our \ours.
\begin{table*}[t]
\centering
\small
\setlength{\tabcolsep}{11pt}
\renewcommand{\arraystretch}{1.15}

\resizebox{0.92\textwidth}{!}{
\begin{tabular}{lcccccc}
\toprule

\multirow{2}{*}{\textbf{Output Length}}
& \multicolumn{6}{c}{\textbf{Prediction Method}} \\

\cmidrule(lr){2-7}

& SSJF-Reg
& SSJF-MC
& LTR-C
& TRAIL
& EGTP
& \cellcolor{blue!8}\textbf{ESTP} \\

\midrule

Short
& 50.37 & \underline{56.70} & 27.56 & 55.82 & 22.03 & \cellcolor{blue!8}\textbf{57.30} \\

Medium
& 80.79 & 70.87 & 62.15 & 66.42 & \underline{81.31} & \cellcolor{blue!8}\textbf{85.44} \\

Long
& 50.75 & 48.21 & 33.20 & 50.48 & \underline{51.00} &  \cellcolor{blue!8}\textbf{52.74} \\

\textbf{Overall}
& \underline{67.67} & 61.60 & 60.46 & 61.15 & 62.36 & \cellcolor{blue!8}\textbf{72.54} \\

\bottomrule
\end{tabular}
}
\caption{
Accuracy across different output length bins in RL scenario. Best results are highlighted in \textbf{bold}, and second-best results are \underline{underlined}. 
ESTP achieves the highest \textbf{overall accuracy} on the Qwen2.5-3B while maintaining competitive performance across Short (<200 tokens), Medium (200–500 tokens), and Long (>500 tokens).
}
\label{tab:acc_bin}
\end{table*}

\paragraph{Experimental Settings.}
We fix the global random seed to 42 to ensure experimental reproducibility and use AdamW~\cite{adam1,adam2} as the optimizer. Hyperparameter $\gamma$ balances entropy and semantic representations, with $\gamma=0.5$. We set $\alpha=1$ to control the strength of prior correction. We set $\lambda=1$ to balance CRPS loss and SmoothL1 loss. $K$ is the number of bins for discretizing the continuous output-length space, with $K=20$. More detailed hyperparameter analysis and experimental settings are provided in the appendix~\ref{app:Hyper} and ~\ref{app:Hyper1}.
\subsection{Main Results}
\paragraph{For Q1: How Does Mean Absolute Error Perform Overall in Length Prediction Tasks? }
We evaluate the Mean Absolute Error across four widely used LLMs, as shown in Table~\ref{tab:main_exp}. Across all four backbones, \ours{} achieves the lowest MAE in every evaluated scenario and consequently obtains the best average result for each model. The improvement is particularly pronounced in the RL scenario, where the average MAE decreases over 20 points. Overall, these results demonstrate that \ours{} consistently improves length prediction across models and scenarios, with especially strong gains for dynamic sampling in RL training.

\begin{figure}[h]
    \centering
    \begin{subfigure}[b]{0.95\linewidth} 
        \centering
        \includegraphics[width=0.95\linewidth]{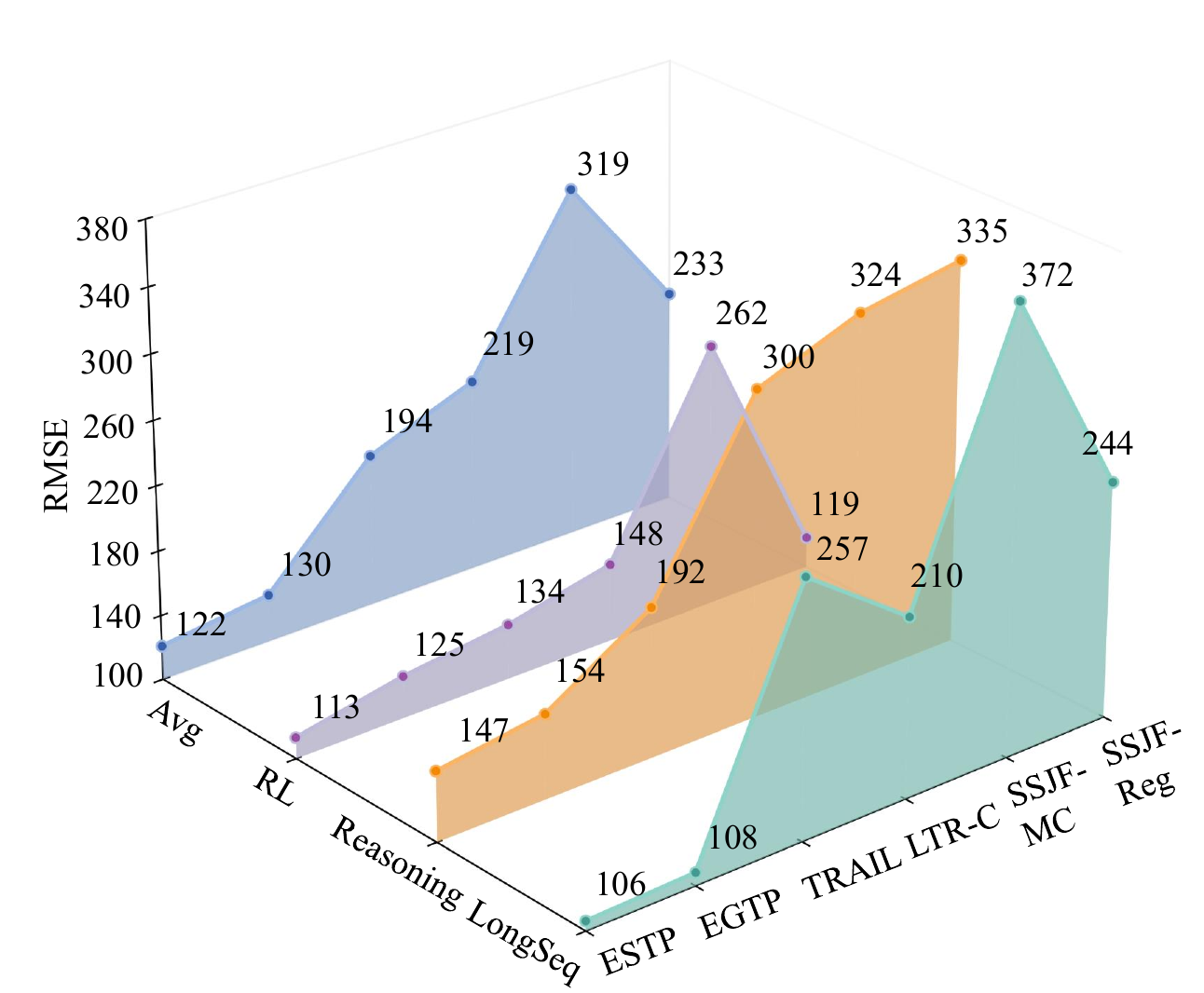}
        \label{fig:RMSE}
    \end{subfigure}
    \caption{Performance comparison of RMSE across different length prediction methods.}
    \label{fig:RMSE}
\end{figure}
\begin{table}[h]
\centering
\setlength{\tabcolsep}{10.0pt}
\scalebox{0.85}{
\begin{tabular}{lcc}
\toprule
\textbf{Method} & \textbf{MAE} & \textbf{RMSE}
\\
\midrule
\textbf{ESTP} & \textbf{929.13$\pm$119.35} & \textbf{1028.57$\pm$131.12}  \\
EGTP & 1012.08$\pm$124.73 & 1183.30$\pm$167.75 \\
\bottomrule
\end{tabular}
}
\caption{Performance on extreme long-CoT regime.}
\label{tab:qwq32b}
\end{table}

\paragraph{For Q2: How Does Root Mean Square Error Penalize Large Deviations in Length Prediction?}
To further evaluate the robustness of our \ours{}, we compare the RMSE values of all approaches on Qwen2.5-7B, which allows us to examine the sensitivity of each method to large prediction errors. As shown in Figure~\ref{fig:RMSE}, \ours{} achieves the best performance in every scenario, and achieves the lowest average RMSE. These results show that our approach provides more accurate and reliable predictions than the baseline methods. We report results on other models in the appendix~\ref{app:more RMSE}.

\begin{table*}[t]
\centering
\small
\setlength{\tabcolsep}{11pt}
\renewcommand{\arraystretch}{1.15}

\resizebox{0.92\textwidth}{!}{
\begin{tabular}{lcccccc}
\toprule

\multirow{2}{*}{\textbf{Metrics}} 
& \multicolumn{6}{c}{\textbf{Prediction Method}} \\

\cmidrule(lr){2-7}

& SSJF-Reg 
& SSJF-MC 
& LTR-C 
& TRAIL 
& EGTP 
& \cellcolor{blue!8}\textbf{ESTP} \\

\midrule

Padding Ratio $\downarrow$   
& 1.34 & 1.32 & 0.59 & 0.91
& \underline{0.52} 
& \cellcolor{blue!8}\textbf{0.46} \\

Throughput $\uparrow$
& 13.21 & 12.48 & 20.22 & 13.30 
& \underline{24.70}
& \cellcolor{blue!8}\textbf{32.75} \\

Avg. JCT $\downarrow$       
& 20.41 & 21.50 & 17.62 & 20.84
& \underline{14.49} 
& \cellcolor{blue!8}\textbf{10.61} \\

\bottomrule
\end{tabular}
}
\caption{
End-to-End System Performance Comparison on Llama3.2-3B. 
Best results are highlighted in \textbf{bold}, and second-best results are \underline{underlined}. 
ESTP achieves the best performance among all methods.
}

\label{tab:duan}
\end{table*}

\paragraph{For Q3: How Does Prediction Accuracy Vary Across Different Length Buckets?}
To better analyze accuracy across different output length bins, we evaluate prediction performance within three intervals (Table~\ref{tab:acc_bin}). \ours{} achieves strong overall accuracy, exceeding 70\% on Qwen2.5-3B, and outperforms TRAIL and EGTP, two state-of-the-art approaches that also leverage the internal activations of LLMs. The SSJF-Reg method achieves competitive performance in terms of overall accuracy, but it relies on an external auxiliary BERT model. In contrast, \ours{} introduces negligible inference latency and memory overhead, making it more efficient and practical than SSJF variants. We report additional results in the appendix~\ref{app:more acc}.
\begin{table}[h]
\centering
\setlength{\tabcolsep}{4.0pt}
\scalebox{0.75}{
\begin{tabular}{lcccc}
\toprule
\textbf{Method} & \textbf{LongSeq} & \textbf{Reasoning} &
\textbf{RL} & \textbf{Average} \\
\midrule
\textbf{ESTP(Ours)} & \textbf{89.47} & \textbf{124.9} & \textbf{73.38} & \textbf{95.92} \\
\midrule
w/o Pos. Debiasing & 95.32 & 138.68 & 78.77 & 104.26 \\
\midrule
w/o Smooth L1 & 102.91 & 138.09 & 80.67 & 107.22 \\
w/o CRPS & 106.05 & 133.07 & 76.45 & 105.19 \\
\midrule
Average Pooling & 163.32 & 133.74 & 92.44 & 129.83 \\
Entropy Pooling & 97.50 & 143.11 & 99.78 & 113.46 \\
Semantic Pooling & 92.79 & 134.69 & 78.53 & 102.00 \\
\bottomrule
\end{tabular}
}
\caption{Ablation Study on the Qwen2.5-3B.}
\label{tab:xiaorong}
\end{table}

\paragraph{For Q4: How much does ESTP outperform the best baseline EGTP in the extreme long-CoT regime?}
We evaluate on the QWQ-LongCoT-AIMO dataset, restricting the analysis to examples with chains of thought longer than 10K tokens. Each method is run independently with five random seeds, and we report the mean and standard deviation across runs. As shown in Table~\ref{tab:qwq32b}, \ours{} outperforms EGTP on this subset, lowering MAE from $1012.08$ to $929.13$ and RMSE from $1183.30$ to $1028.57$. The larger reduction in RMSE indicates that \ours{} more effectively mitigates large prediction errors, demonstrating stronger robustness on exceptionally long reasoning sequences.

\subsection{Ablation Analysis}
As Table~\ref{tab:xiaorong} shows, \ours{} achieves the lowest average MAE of 95.92 across three scenarios. Removing position debiasing increases the average MAE to 104.26, confirming the importance of correcting positional bias in attention scores. Excluding Smooth L1 or CRPS degrades performance, indicating that both loss terms improve length prediction accuracy. Replacing entropy-semantic pooling with average, entropy-only, or semantic-only pooling consistently degrades performance. These results show that entropy and semantic information are complementary and that each \ours{} component contributes to overall performance.

\begin{figure}[t]
    \centering
    \begin{subfigure}[b]{0.95\linewidth} 
        \centering
     \includegraphics[width=0.95\linewidth]{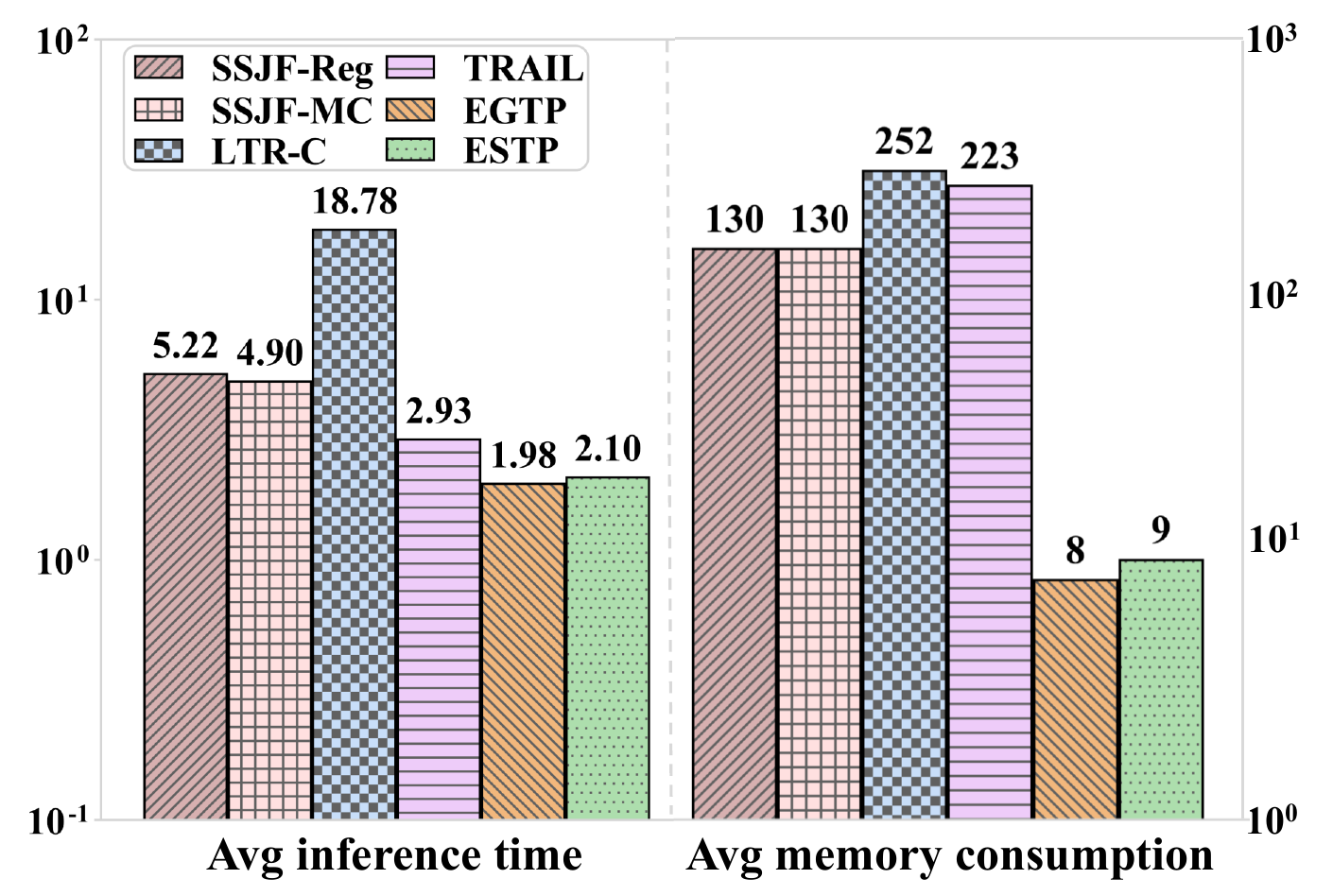}
        \label{fig:time&mem}
    \end{subfigure}
    \caption{Average time(ms) and memory(MB) overhead per sample for predicting in LongSeq scenario.}
    \label{fig:time&mem}
\end{figure}

\subsection{End-to-End System Performance}
We integrated \ours{} and baselines with the Shortest Job First (SJF) scheduler in an end-to-end system backed by the vLLM serving engine. \ours{} achieves the best performance in the RL setting(Table~\ref{tab:duan}). Compared with EGTP, it reduces the padding ratio and increases throughput by over 30\%. In summary, \ours{} serves as a practical, effective building block for length-aware LLM serving systems. It enables better resource allocation and scheduling to boost system throughput.

\subsection{Efficiency Analysis}
We provide a cost comparison of \ours{} with baselines in Figure~\ref{fig:time&mem}. Avg inference time represents the average time to process a query and generate the prediction. On LongSeq, \ours{} from entropy computation and attention-based semantic weighting to final prediction, takes 2.10 ms per sample, and \ours{} has much lower inference latency than methods that use external auxiliary models like SSJF (BERT) and LTR-C (OPT-350M), since these models bring significant computational overhead and longer response time. For memory consumption, our \ours{} based on prefill features also use far less memory than those relying on external auxiliary models. Overall, \ours{} achieves a favorable trade-off between average inference time and memory footprint, and performs well on both dimensions. 

\section{Conclusion}
In this paper, we first point out the major drawback of entropy-only guided pooling: it does not reflect the semantic significance of individual tokens well. To address this problem, we present \ours{}, a lightweight framework that combines attention-based semantic representations with token-level entropy and uses an ordinal length-distribution prediction head to cope with the heavy-tailed nature of generation lengths. Experiments on several tasks show that our approach achieves better results than prior methods across most scenarios, and it also brings considerable improvements on extreme long-CoT regime and in end-to-end system performance. 

\section*{Limitations}
The limitations of our work reside in three primary aspects.
First, \ours{} depends on the accessible hidden states, per-head attention scores, and token-level predictive distributions derived from the prefill stage of open-source autoregressive LLMs. Since closed-source commercial LLM platforms do not expose intermediate internal activations and token-level probability distributions, our method cannot be directly applied to black-box proprietary models, which limits its compatibility and deployment scope.
Second, in this work, we adopt debiased attention as a useful proxy for semantic importance. A set of validation experiments demonstrate its consistency with semantic importance. Nevertheless, debiased attention cannot fully substitute semantic importance. Therefore, we leave for future work the exploration of better features that can serve as proxies for semantic importance.
Third, our evaluation relies on a limited set of datasets and does not cover other extreme and complex scenarios. We leave validating the effectiveness of our method across broader datasets and complex settings for future work.


\section*{Ethics Statement}
Our research focuses on accurate LLM output length prediction and lightweight serving optimization, aiming to reduce padding overhead and improve hardware utilization for real-world LLM deployment. In this study, we explicitly elaborate on three core ethical considerations inherent to our research.

First, all datasets and baseline models adopted in this research are publicly accessible. No private, confidential or personally sensitive information is utilized during model training, validation and testing phases. All experimental procedures strictly abide by relevant data governance norms, effectively preventing privacy disclosure and potential ethical hazards.

Second, we will release the implementation code and experimental configurations for public access to facilitate reproducible research and further studies on LLM length prediction and serving optimization. Our method only optimizes inference scheduling and output length prediction without modifying the original LLM’s inherent generation logic, safety alignment, and decision mechanism, thus introducing no bias amplification or potential safety risks.

Finally, during drafting and revision, we used AI assistants to help optimize writing and improve clarity. No content was generated unsupervised or without verification.


\bibliography{custom}

\appendix

\clearpage

\section{Evaluation Metrics}  
\label{app:metrics}

\subsection{Mean Absolute Error}
\label{app:MAE}
To precisely assess the prediction accuracy of our ESTP, we utilize Mean Absolute Error (MAE) as a primary metric.Mean Absolute Error (MAE) measures the average absolute difference between the predicted values ($\hat{y}_{i}$) of the prediction head and the ground-truth values ($y_{i}$). A lower MAE indicates more accurate prediction results of the prediction head. For a given set of 
$n$ samples, it is defined as:
\begin{equation}
\label{eq:1-1}
MAE=\frac{1}{n}\sum_{i=1}^{n}\left | y_{i}- \hat{y}_{i}  \right |   
\end{equation}

\subsection{Root Mean Square Error}
\label{app:RMSE}
To precisely assess the prediction accuracy of our ESTP, we utilize Root Mean Squared Error (RMSE) as a primary metric.
Root Mean Squared Error (RMSE) measures the root mean squared difference between the predicted values ($\hat{y}_{i}$) of the prediction head and the ground-truth values ($y_{i}$). A lower RMSE indicates more accurate prediction results of the prediction head. For a given set of $n$ samples, it is defined as:
\begin{equation}
\label{eq:1-2}
RMSE=\sqrt{\frac{1}{n}  {\textstyle \sum_{i=1}^{n}}(\hat{y}_{i}-y_{i})^{2}} 
\end{equation}

\subsection{Accuracy across different output length bins}
\label{app:RMSE}
We define a binning-based accuracy to further evaluate prediction reliability.
Given interval bins [0, 200, 500], each ground-truth value $y_i$ and predicted value $\hat{y}_i$
are mapped to corresponding bins. The bin accuracy is calculated as the ratio of samples
where the prediction bin matches the ground-truth bin.
A higher bin accuracy indicates more reliable and robust prediction. For a given set of samples, it is defined as:
\begin{equation}
\label{eq:1-3}
Acc_{bin}=\frac{N_{match}}{N_{total}} 
\end{equation}
where $N_{match}$ denotes the number of samples whose predicted bins are consistent with the ground-truth bins, and $N_{total}$ denotes the total number of test samples.

\subsection{Throughput}
\label{app:throughput}
We use throughput to measure the number of tasks successfully processed by the system per unit of time. The specific formula is as follows, where $T_{total}$ represents the time interval, and $N_{completed}$ represents the amount of tasks successfully completed during that period.
\begin{equation}
\label{eq:1-4}
Throughput=\frac{N_{completed} }{T_{total} } 
\end{equation}

\subsection{JCT}
\label{app:jct}
We use average JCT to measure the mean completion time of all tasks (from task submission to task completion). The specific formula is as follows:
\begin{equation}
\label{eq:1-5}
JCT=T_{completion}-T_{submission}  
\end{equation}

\subsection{Padding Ratio}
\label{app:pad}
We use Padding Ratio to quantify the overhead or resource waste caused by padding. The specific formula is as follows, where $R_{actual}$ is the actual required resource size, and $R_{allocated}$ is the total resources allocated by the system:
\begin{equation}
\label{eq:1-6}
Padding\ Ratio=\frac{R_{allocated}-R_{actual}  }{R_{actual} } 
\end{equation}
\begin{table}[th]
\centering
\small
\setlength{\tabcolsep}{2pt}
\renewcommand{\arraystretch}{1.15}

\resizebox{0.45\textwidth}{!}{
\begin{tabular}{lc}
\toprule

\multirow{1}{*}{\textbf{Splitting strategy}}
& QWQ-LongCoT-AIMO($>$10k) 
\\
\midrule
Train
& 269\\
Validation
& 30\\
Test
& 36\\

\bottomrule
\end{tabular}
}
\caption{
Dataset of QWQ-LongCoT-AIMO($>$10k).
}
\label{tab:dataset2}
\end{table}

\begin{table}[h]
\centering
\small
\begin{tabular}{c c c c c}
\toprule
\multirow{1}{*}{
\textbf{Value}} & \textbf{Reasoning} & \textbf{RL} & \textbf{LongSeq} & \textbf{Avg} \\
\midrule
0.3 & 109.03 & 61.63 & 94.74 & 88.47 \\
0.4 & 108.53 & 59.58 & 94.01 & 87.37\\
\textbf{0.5} & \textbf{108.45} & \textbf{58.66} & \textbf{93.35} & \textbf{86.82}\\
0.6 & 109.55 & 58.83 & 93.75 & 87.38 \\
0.7  & 109.59 & 59.33 & 93.40 & 87.44 \\
\bottomrule
\end{tabular}
\caption{Hyperparameter Sensitivity Analysis.}
\label{tab:Par}
\end{table}
\begin{table*}[htbp]
\centering
\small
\setlength{\tabcolsep}{11pt}
\renewcommand{\arraystretch}{1.15}

\resizebox{0.92\textwidth}{!}{
\begin{tabular}{lcccccc}
\toprule

\multirow{2}{*}{\textbf{Splitting strategy}}
& \multicolumn{3}{c}{\textbf{Qwen2.5\_3B}} 
& \multicolumn{3}{c}{\textbf{Qwen2.5\_7B}} 
\\

\cmidrule(lr){2-4}
\cmidrule(lr){5-7}
& Reasoning
& RL
& LongSeq
& Reasoning
& RL
& LongSeq \\

\midrule

Train
& 324 & 50253 & 660 & 324 & 52578 & 660 \\
Validation
& 104 & 16751 & 220 & 106 & 17527 & 177 \\
Test
& 113 & 16255 & 177 & 108 & 17213 & 220 \\

\bottomrule
\end{tabular}
}
\caption{
Dataset splitting strategy on the Qwen2.5.
}
\label{tab:dataset}
\end{table*}
\begin{table*}[htbp]
\centering
\small
\setlength{\tabcolsep}{11pt}
\renewcommand{\arraystretch}{1.15}

\resizebox{0.92\textwidth}{!}{
\begin{tabular}{lcccccc}
\toprule

\multirow{2}{*}{\textbf{Splitting strategy}}
& \multicolumn{3}{c}{\textbf{llama3.2\_1B}} 
& \multicolumn{3}{c}{\textbf{llama3.2\_3B}} 
\\

\cmidrule(lr){2-4}
\cmidrule(lr){5-7}
& Reasoning
& RL
& LongSeq
& Reasoning
& RL
& LongSeq \\

\midrule

Train
& 299 & 54496 & 660 & 295 & 52998 & 660 \\
Validation
& 90 & 18166 & 220 & 97 & 17515 & 220 \\
Test
& 152 & 17825 & 42 & 149 & 17819 & 32 \\

\bottomrule
\end{tabular}
}
\caption{
Dataset splitting strategy on the Llama3.2.
}
\label{tab:dataset1}
\end{table*}
\section{Dataset Details}
\label{app:dataset split}
We present the dataset split sizes for each scenario under every backbone model in Table~\ref{tab:dataset} and Table~\ref{tab:dataset1}. Meanwhile, Table~\ref{tab:dataset2} lists the split sizes with more than 10k tokens samples in the QWQ-LongCoT-AIMO dataset.

\begin{table*}[h]
\centering
\setlength{\tabcolsep}{6pt}
\renewcommand{\arraystretch}{1.15}

\resizebox{0.95\textwidth}{!}{
\begin{tabular}{lcccc}
\toprule

\textbf{Reference and metric}
& Raw mean
& Raw median
& Debiased mean
& Debiased median
\\

\midrule

Causal baseline  Pearson
& 0.5401 & 0.5222 & 0.0851 & 0.1306 \\

Causal baseline  Spearman
& 0.5462 & 0.5692 & 0.0861 & 0.1103 \\

Normalized position  Pearson
& -0.2633 & -0.2449 & -0.0892 & -0.1152 \\

Normalized position  Spearman
& -0.5462& -0.5692& -0.0861 & -0.1103 
\\
\bottomrule
\end{tabular}
}
\caption{
Correlation of token scores with position based references.
}
\label{tab:prior}
\end{table*}

\section{Analysis of Position Debiasing}
\label{app:prior}
Table~\ref{tab:prior} shows that debiasing substantially weakens the dependence of incoming attention scores on token position. The mean Pearson and Spearman correlations with the causal position baseline fall from 0.5401 and 0.5462 to 0.0851 and 0.0861, respectively, with similar reductions in the median correlations. The absolute correlations with normalized token position also decrease. Raw incoming attention is therefore strongly shaped by the greater causal visibility of earlier tokens, whereas debiasing removes most of this position-related signal. The corrected score is thus more suitable for semantic token weighting because it is less dominated by a token's position in the prompt.

\begin{table*}[t]
\centering
\setlength{\tabcolsep}{6pt}
\begin{tabular*}{0.95\textwidth}{@{\extracolsep{\fill}}llcccccc}
\toprule

\multirow{2}{*}{\textbf{Model}} 
& \multirow{2}{*}{\textbf{Input Length}}
& \multicolumn{6}{c}{\textbf{Prediction Method}} \\

\cmidrule(lr){3-8}

&
& SSJF-Reg
& SSJF-MC
& LTR-C
& TRAIL
& EGTP
& \textbf{ESTP} \\

\midrule

\multirow{4}{*}{\textbf{Llama3.2-1B}}

& Short
& 72.53 & 67.21 & \underline{73.59} & 68.20 & 28.61 & \textbf{76.41} \\

& Medium
& 60.10 & 63.12 & 46.81 & 52.63 & \textbf{79.50} & \underline{66.11} \\

& Long
& 55.08 & 58.41 & 14.80 & \underline{60.54} & 55.67 &\textbf{62.16} \\

& \textbf{Overall}
& 66.13 & 65.98 & \underline{66.64} & 59.68 & 46.54 & \textbf{72.09} \\

\midrule

\multirow{4}{*}{\textbf{Llama3.2-3B}}

& Short
& 60.31 & 61.16 & \textbf{77.54} & 51.22 & 24.10 & \underline{62.91} \\

& Medium
& 71.82 & 65.48 & 57.72 & 69.26 & \underline{74.18} & \textbf{75.18} \\

& Long
& 41.27 & \textbf{64.19} & 27.66 & \underline{48.32} & 27.66 & 43.04 \\

& \textbf{Overall}
& 62.82 & \underline{65.59} & 64.27 & 61.25 & 43.67 & \textbf{66.74} \\

\bottomrule
\end{tabular*}

\caption{
Accuracy comparison across different output length bins in RL scenarios(\textbf{higher is better}). 
}

\label{tab:acc_appendix}
\end{table*}

\begin{table*}[t]
\centering
\setlength{\tabcolsep}{11pt}
\begin{tabular}{l l | ccccc }
\toprule
\multirow{2}{*}{\textbf{Model}} & \multirow{2}{*}{\textbf{Scenario}} & \multicolumn{5}{c}{\textbf{Prediction Method}\vspace{1mm}}  \\
& & SSJF-Reg & SSJF-MC & LTR-C & EGTP & \textbf{ESTP} \\
\midrule
\multirow{4}{*}{\textbf{Qwen2.5 3B}} & LongSeq
& 302.82 & 351.14 & 210.61 & 109.42 & \textbf{103.07}
\\
& Reasoning & 363.15 & 291 & 329.58 & 170.38 & \textbf{145.72}\\
& RL & 109.33 & 247.79 & 130.94 &  116.6 & \textbf{104.52}\\
& \textbf{Avg} & 258.43 & 296.64 & 223.71 & 132.13 & \textbf{117.77}\\
\midrule 
\multirow{4}{*}{\textbf{Llama3.2 1B}} & LongSeq
& 448.9 & 215.84 & 410.37 &  104.41 & \textbf{103.02} \\
& Reasoning & 363.38 & 264.04 & 341.92 &  198.52 & \textbf{195.35} \\
& RL & 118.21 & 228 & 189.25 & 140.41 & \textbf{112.73} \\
& \textbf{Avg} & 310.16 & 235.96 & 313.85 &  147.78 & \textbf{137.03} \\
\midrule 
\multirow{4}{*}{\textbf{Llama3.2 3B}} & LongSeq
& 438.58 & 230.51 & 418.22 & \textbf{126.30} & 148.27 \\
& Reasoning & 383.91 & 241.88 & 360.44 &  164.21 & \textbf{159.60}\\
& RL & 162.25 & 247.14 & 204.61 &  184.53 & \textbf{159.48} \\
& \textbf{Avg} & 328.25 & 239.84 & 327.76 & 158.35 & \textbf{155.78} \\
\bottomrule
\end{tabular}
\caption{Comparison of RMSE Metrics Across Different Length Prediction Methods.}
\label{tab:rmse}
\end{table*}
\clearpage
\begin{figure*}[t]
    \centering
    \begin{subfigure}[b]{1.0\linewidth}
        \centering
        \includegraphics[width=1.0\linewidth]{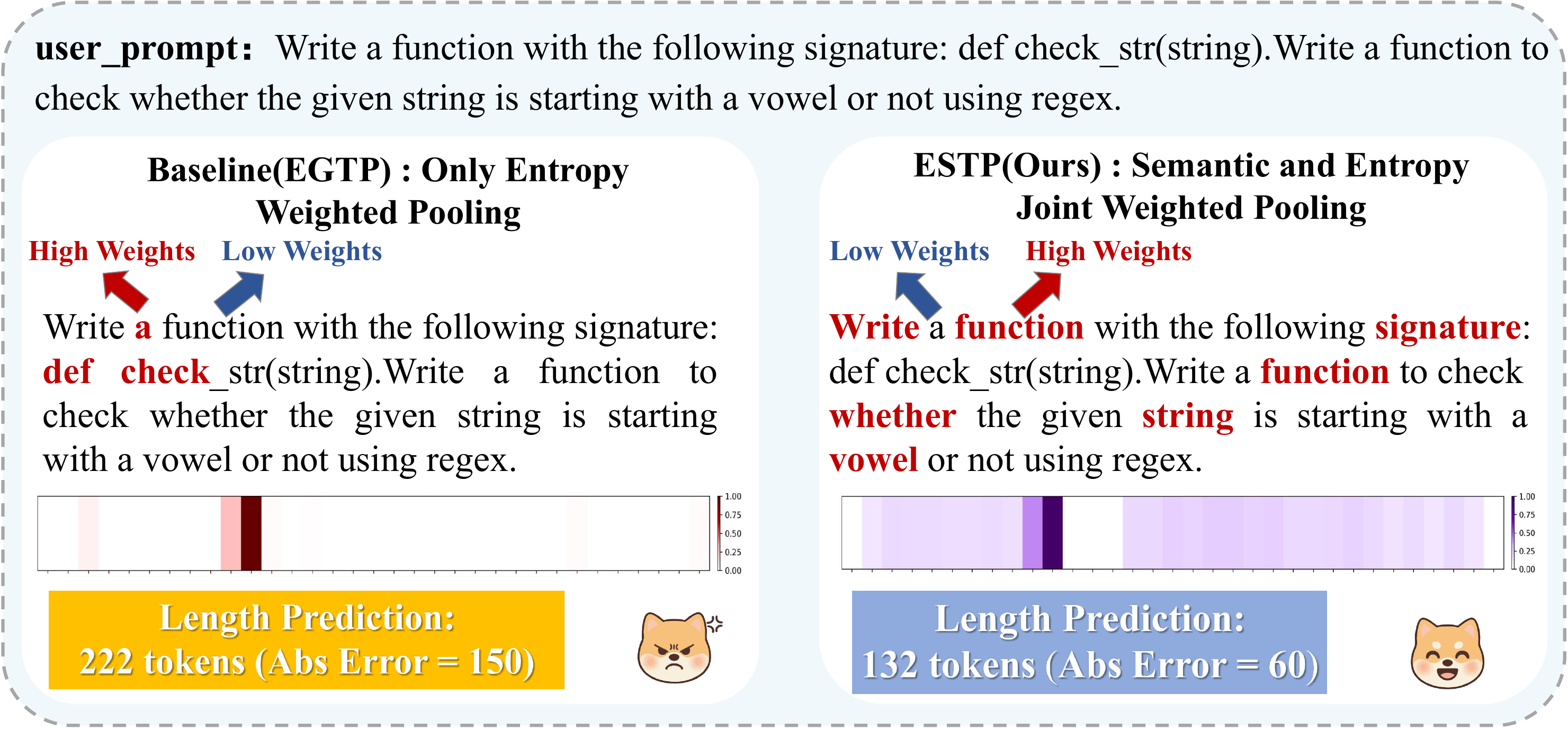}
        \label{fig:case}
    \end{subfigure}
    \caption{Qualitative comparison demonstrates the token weighting distributions of EGTP and ESTP on specific samples. Red tokens indicate high-weight tokens, and the corresponding prediction results are provided for reference.}
    \label{fig:case}
\end{figure*}
\section{More Experiments}
\label{app:more exp}
We provide the code, experimental hyperparameter settings and the dataset in the supplementary materials.
\subsection{RMSE on the other Models} 
\label{app:more RMSE}
To demonstrate the generalization and model-agnostic property of our proposed ESTP framework, we further evaluate the critical metric RMSE on Qwen2.5-3B, Llama3.2-1B, and Llama3.2-3B models. As shown in Table~\ref{tab:rmse}, our approach achieves superior prediction performance across all evaluated models.
Specifically, our ESTP method achieves the best prediction performance in both the Reasoning and RL scenarios. Meanwhile, on the Qwen2.5-3B model, it obtains an average RMSE of only 117.77, which is lower than that of the SSJF-series methods and the LTR-C method. These experimental results fully demonstrate that our ESTP approach possesses stronger model generalization and more accurate prediction.

\subsection{Accuracy comparison on the other Models} 
\label{app:more acc}
To better analyze accuracy across different length bins, we evaluate prediction performance on the Llama3.2-1B/3B. As shown in Table~\ref{tab:acc_appendix}. Our ESTP still maintains strong predictive performance across different bins while also achieving the best Overall Accuracy on Llama3.2-1B/3B.

\subsection{Hyperparameter Sensitivity Analysis}
\label{app:Hyper}
In this section, we provide an in-depth analysis of the impact of the key hyperparameter : the $\gamma $ to balance the entropy representation and semantic representaion. We conduct our study on the Qwen2.5-3B model to investigate how adjustments to this parameter affect the final model performance on output length prediction.The empirical results are summarized in Table~\ref{tab:Par}. The performance peaks at $\gamma=0.5 $, yielding the best results across all scenarios.



\subsection{Other Hyperparameter Setting} 
\label{app:Hyper1}
We set the learning rate of 2e-5. 
In the end-to-end experiments, We evaluate 1,000 requests under a closed-batch workload in which all requests are available at the beginning of the SJF evaluation. 

\section{Normalized Smooth L1 Loss}
\label{app:smooth}
Let $L_{\min}$ and $L_{\max}$ denote the 1st and 99th percentiles of the output lengths in the training set, respectively. For sample $n$, the predicted length is computed as the expectation over the predicted length distribution:
\begin{equation}
\hat{y}_n = \sum_{k=1}^{K} \hat{p}_{n,k} c_k,
\end{equation}
where $\hat{p}_{n,k}$ is the predicted probability of the $k$-th length bin and $c_k$ is its center. We normalize the predicted and ground-truth lengths as
\begin{equation}
\tilde{\hat{y}}_n =
\frac{\hat{y}_n-L_{\min}}{L_{\max}-L_{\min}},
\qquad
\tilde{y}_n =
\frac{y_n-L_{\min}}{L_{\max}-L_{\min}}.
\end{equation}
Let $d_n=\tilde{\hat{y}}_n-\tilde{y}_n$. The normalized Smooth L1 loss over a batch of $B$ samples is
\begin{equation}
\mathcal{L}_{\mathrm{SmoothL1}}
=
\frac{1}{B}\sum_{n=1}^{B}\ell_{\beta}(d_n),
\end{equation}
where
\begin{equation}
\ell_{\beta}(d)=
\begin{cases}
\dfrac{d^2}{2\beta}, & |d|<\beta, \\[4pt]
|d|-\dfrac{\beta}{2}, & \text{otherwise}.
\end{cases}
\end{equation}
We use $\beta=1$ and do not clip the ground-truth lengths after normalization. The quadratic region provides smooth gradients for small prediction errors, while the linear region reduces sensitivity to large errors. The complete training objective is
\begin{equation}
\mathcal{L}
=
\mathcal{L}_{\mathrm{CRPS}}
+
\lambda\mathcal{L}_{\mathrm{SmoothL1}},
\end{equation}
where $\lambda=1$ in our experiments.

\section{Case Study}
\label{app:case_study}
We visualize token-level weights on real examples to show the advantage of our \ours{} method, which combines entropy-based signals with semantic features. The results illustrate that our model does a better job at capturing the core meaning of the input prompts. In Figure~\ref{fig:case}, we see that our method puts more weight on important words and entities, which helps extract key information more precisely. In contrast, EGTP amplifies the weights of tokens that have little semantics, while underweighting those carry important information. For final length prediction, \ours{} with combined entropy and attention-based semantic information achieves significantly lower prediction errors than EGTP only with entropy.

\end{document}